\documentclass[10pt,journal]{IEEEtran}

\usepackage{graphicx}
\usepackage{subfigure}
\usepackage[hyphens]{url}
\usepackage{hyperref}  
\usepackage{xcolor}

\begin{document}

\title{Intelligence-Endogenous Management Platform for Computing and Network Convergence}

\author{Zicong~Hong,~\IEEEmembership{Graduate Student Member,~IEEE,}
        Xiaoyu~Qiu,
        Jian~Lin, 
        Wuhui~Chen,~\IEEEmembership{Member,~IEEE,}
        Yue~Yu, 
        Hui~Wang, 
        Song~Guo,~\IEEEmembership{Fellow,~IEEE,}
        and Wen~Gao,~\IEEEmembership{Fellow,~IEEE}}

\maketitle

%TC:ignore
\begin{abstract}
Massive emerging applications are driving demand for the ubiquitous deployment of computing power today. 
This trend not only spurs the recent popularity of the \emph{Computing and Network Convergence} (CNC), but also introduces an urgent need for the intelligentization of a management platform to coordinate changing resources and tasks in the CNC.
Therefore, in this article, we present the concept of an intelligence-endogenous management platform for CNCs called \emph{CNC brain} based on artificial intelligence technologies.
It aims at efficiently and automatically matching the supply and demand with high heterogeneity in a CNC via four key building blocks, i.e., perception, scheduling, adaptation, and governance, throughout the CNC's life cycle.
Their functionalities, goals, and challenges are presented.
To examine the effectiveness of the proposed concept and framework, we also implement a prototype for the CNC brain based on a deep reinforcement learning technology.
Also, it is evaluated on a CNC testbed that integrates two open-source and popular frameworks (OpenFaas and Kubernetes) and a real-world business dataset provided by Microsoft Azure.
The evaluation results prove the proposed method's effectiveness in terms of resource utilization and performance.
Finally, we highlight the future research directions of the CNC brain.
\end{abstract}
%TC:endignore

\section{Introduction}
\label{sec:introduction}

With the development of emerging applications such as the metaverse, AI chatbots and autonomous driving, computing power has become the most important and innovative form of productivity \cite{zhu2022space,smolensky2022neurocompositional}. 
Specifically for the metaverse, the computing power required to deliver a virtual world to each participant would be significant, as the system would need not only to track massive objects, characters and environmental effects, but also to adapt the display as any or all of these move through the virtual space for interaction, immersion and imagination \cite{wang2022survey,lim2022realizing}. 
In the future, an ideal metaverse is expected to accommodate millions or even billions of users, which Intel predicts will require a thousandfold increase in computing power~\cite{intel}.

Over the past decade, as the demand for computing power has increased, cloud computing has become popular in academia and industry \cite{mehonic2022brain}. Users can outsource their computing tasks to data centers with large computing power provided by cloud providers. A complementary approach is edge computing~\cite{10.1145/3464419}, where computing power is placed at the edges of the network, such as base stations or network gateways, to improve response times and save bandwidth. Today, many data centers or edge nodes have been built in the network, and by 2023, the global computing power will reach a 500-fold increase compared to 2020~\cite{huawei_giv}.

However, the growing computing needs of emerging applications and the new deployment of computing nodes are unevenly distributed regionally. 
This is because densely populated and economically active regions have more users and computing tasks, while resource-rich and sparsely populated regions have lower costs for deploying, operating and maintaining computing nodes. 
In addition to location differences, the hardware heterogeneity of computing nodes and the objective diversity of emerging applications complicate the distribution of supply and demand. 
Ultimately, this creates an imbalance between the supply and demand of computing power, preventing it from being fully utilised and compromising the quality of service of emerging applications. 
% 
% Most of the emerging applications' computation requires the cooperation of multiple providers.
% For example, ...
% However, in practice, there is ...
% The problem is called \emph{computing power silos}.

% Considering the limitation of Moore's Law and the expense of building new data centres, 
To solve the problem, it has become a growing direction to connect various computing nodes (cloud servers, edge servers, and PCs) distributed in the network, and improve the overall scale of computing power.
It thus introduces a new concept named \emph{Computing and Network Convergence} (CNC).
The CNC is expected to provide easy-to-use, high-quality, and ubiquitous computing power in the network, enabling kinds of emerging applications to utilize computing power in the way we use electricity or water today.

Despite the benefits and progress made towards this goal, there is still much to be done. 
One of the major challenges is to coordinate resources and tasks distributed in the CNC so that users can easily access the right resources for their different needs. 
Different business scenarios require different levels of computing power, and the same customer may require multiple types of computing power. 
In terms of types of collaboration, this includes resource planning across clouds, edges and ends, as well as across industries, regions and people.
This drives demand for a higher level of efficient resource management and intelligent optimizations for CNCs.

In this article, we propose the concept of an intelligence-endogenous management platform for CNCs, called \emph{CNC brain}.
Rather than relying on many professional works, the CNC brain utilizes artificial intelligence technologies to manage a CNC with unprecedented geographical scale and operation complexity.
A CNC brain should support four key functionalities, i.e., \emph{perception}, \emph{scheduling}, \emph{adaptation}, and \emph{governance}.
In particular, perception is the prerequisite for management, which aims to quickly and comprehensively obtain and analyse the distribution and status of a variety of resources and tasks across the CNC.
Depending on the results of perception, scheduling aims for flexible, adaptive, and holistic task allocation and resource scheduling to match supply and demand in the CNC.
During the scheduling, the complicated, dynamic environment of the CNC requires a  design of adaptation to help the manager be resilient to failures for reliable computing services.
Last but not least, massive participants and pluralistic interests behind them in CNC are calling for governance that reaps the benefits of the CNC while resolving the growing concerns about property rights, social risk, and resource marketing in the CNC.

Following the above framework design, we implement a prototype of the CNC brain based on deep reinforcement learning (DRL) technology over a CNC testbed integrating OpenFaas (one of the most popular serverless frameworks) and Kubernetes (an open-source platform for managing containerized applications).
The results of evaluating it under a real-world business dataset publicly released by Microsoft Azure show its resource utilisation efficiency and low service-level agreement (SLA) violation rate.

% The main contributions of this article are summarized as follows:
% \begin{itemize}
%     \item ...
%     \item ...
%     \item Various experiments are conducted to verify the effectiveness of the proposed method in both aspects of resource usage and performance.
% \end{itemize}

In what follows, this article first offers an overview of the state-of-the-art CNCs and their general characteristics and provides an application scenario. 
Then, it introduces a CNC brain framework and its four key functionalities. 
After that, DRL-based intelligent resource management is proposed as a prototype of the CNC brain and evaluated. 
Finally, the article is concluded with technical concerns and future challenges.

\section{Background: Computing and Network Convergence}
\label{sec:background}

\subsection{Existing Works}

Recently, there have been several CNCs worldwide, for which we introduce the design, plan and vision of three representatives as follows.

\paragraph{China Computing NET} Currently, the total scale of computing power in China has exceeded 140 EFLOPS, with an average annual growth rate of over 30\% in the past five years, making it the second-largest scale of computing power in the world. 
Nevertheless, with the explosive growth of various smart scenarios (e.g., AI, big data, IoT, blockchain), the computing power demand is also growing dramatically (especially in the eastern region). 
``Channel computing resources from east to west'' is a strategic project for the cross-region deployment of computing resources to address the imbalance between the supply and demand of computing resources in the east and west of China. 
Specifically, the project hopes to leverage the energy advantages of the central and western regions to build computing infrastructure (plan to build 8 national computing power center nodes and ten national data center clusters) to serve the computing power deficit regions such as the east coast.  
However, the project requires data transmission across regions, significantly increasing data transmission latency. Therefore, a flexible scheduling mechanism for computing resources (including general-purpose, intelligent, supercomputing, and edge computing power) is urgently needed.

\paragraph{Sky Computing} Existing multiple cloud service providers offer proprietary interfaces, which inconveniences customers (e.g. customers are locked into a particular provider, even if other cloud platforms offer more affordable options or better services). Sky computing is a promising solution to help customers place workloads flexibly. Sky computing refers to an inter-cloud broker that helps individual customers select the appropriate cloud platform for their workload, and customers can rely on the broker to optimise their desired criteria (e.g. price, performance) \cite{chasins2022sky}. In addition to customer benefits, sky computing gives third-party software services companies a greater competitive advantage. This is because greater cloud compatibility will make it easier for these third-party software services to be ported to multiple clouds, allowing them to reach additional customers.

\paragraph{Gensyn.ai} The Gensyn network\footnote{https://www.gensyn.ai/} is one of the most popular distributed protocols for machine learning that aims to unite the computing power distributed in the world into a global CNC that is accessible to anyone at any time. Various computing devices can easily connect to the protocol, making the computing power accessible to engineers, researchers, and academics with AI training needs.

\subsection{Characteristics}

While there are a few implementations for the CNCs, they have three general characteristics.

\textit{Characteristics 1: Serverless provision.} Recently, an emerging cloud computing paradigm, serverless computing, has gained widespread attention due to its characteristics of autonomous resource scalability, ease of use, and pay-as-you-go charging model \cite{ali2022optimizing}. More specifically, serverless technology facilitates the deployment of diverse applications (e.g., machine learning, data analytics) by enabling rapid application deployment, pay-as-you-go, and seamless application scaling without the need to manage complex computing resources. Despite the many conveniences serverless computing offers, there are still challenges that hinder its further development, including fine-grained auto-scaling strategies, effective cold-start management and selecting the right serverless provider.

\textit{Characteristics 2: Heterogeneous resources computing.} CNC will be a new evolution of multi-access cloud/edge computing, which is expected to flexibly utilise heterogeneous computing resources (e.g. CPU, GPU, FPGA and ASIC) provided by different cloud or edge servers. Considering the respective advantages of different computing resources compared to users directly using a particular computing resource, the heterogeneous computing resources can work together to accelerate operations (e.g., kernels or tasks in an AI application) \cite{zhao2022multi}. However, it is challenging to integrate the heterogeneous computing resources across different cloud and edge servers in the CNC for workloads because different cloud or edge servers have different hardware characteristics (e.g., instruction set architectures, frequencies, cache sizes, and I/O bandwidth) and different software characteristics (e.g., computing platform and application programming interface). 

\textit{Characteristics 3: Computing-network integration.} In-network computing refers to a special type of hardware acceleration where network traffic is intercepted and computational tasks are performed by network devices before reaching the host \cite{lao2021atp}. Therefore, the design can effectively improve the quality of service (e.g. response latency). However, further exploitation of in-network resources to improve the quality of service is challenging due to network packet loss, highly heterogeneous programming models of in-network hardware, non-linear sharing of in-network resources, and fine-grained locality constraints regarding the underlying network topology. 

\begin{figure}[t]
    \centering
    \includegraphics[width=\linewidth]{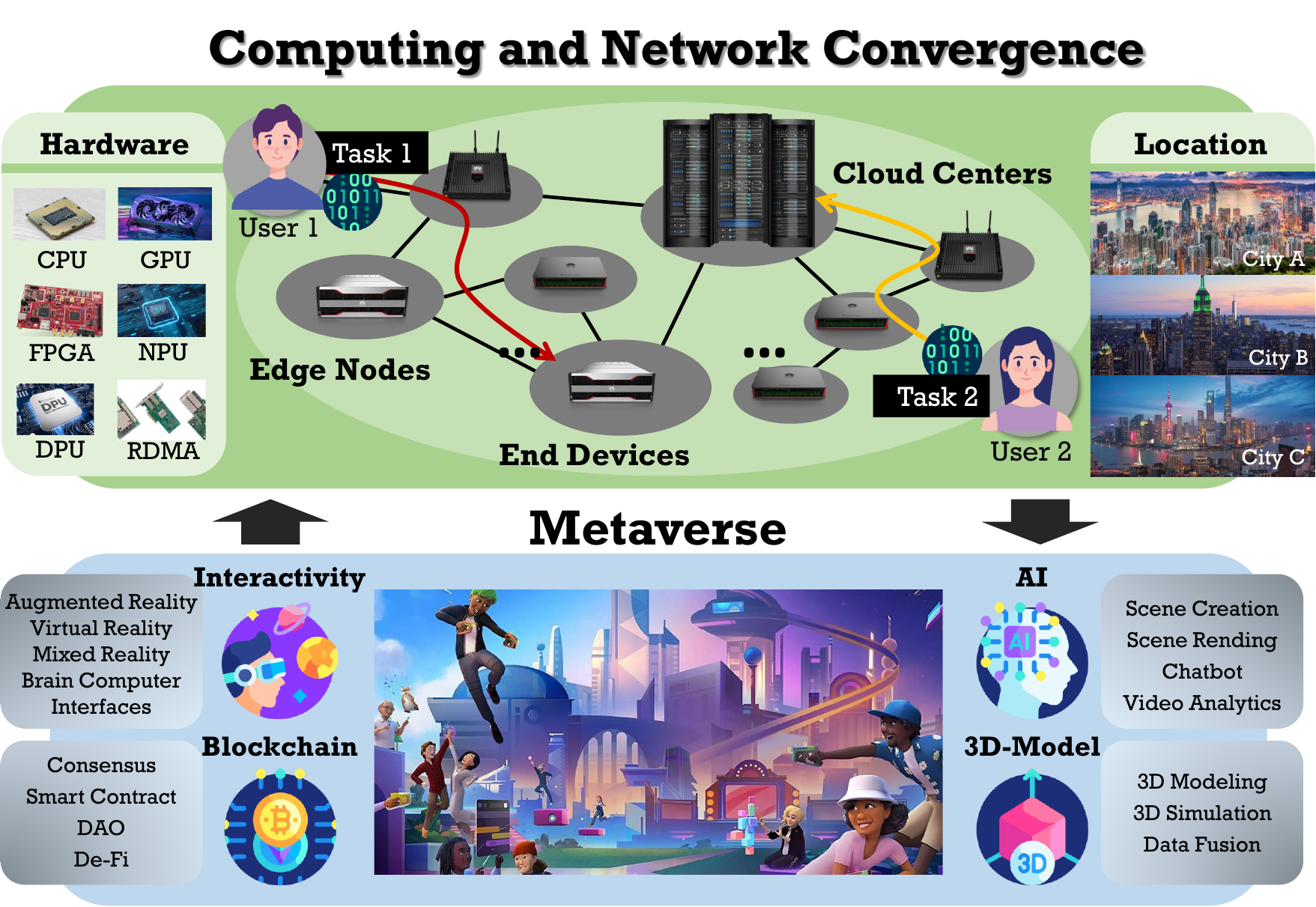}
    \caption{Application scenario for CNC in the metaverse.}
    \label{fig:Background}
\end{figure}

\subsection{Application Scenario}

Fig. \ref{fig:Background} shows how a CNC can empower the applications in the metaverse. 
CNC provides the appropriate computing resource allocation strategy and fast, low-latency connectivity that is essential for real-time interaction and immersive experiences in the metaverse. With CNC, users can move seamlessly between the virtual and physical worlds with little to no experiential lag or delay. In addition, users can also get a consistent experience regardless of where they are and what device they are using.
Specifically, for latency-sensitive and non-computation-intensive tasks (e.g., pose recognition), CNC can intelligently assign that computational task to the nearest edge computing node1.
For latency-insensitive and computationally intensive tasks (e.g., applying empirically enhanced training to online data), CNC can assign that task to a relatively distant cloud computing center.

\section{Computing and Network Convergence Brain Framework}

\begin{figure*}[t]
    \centering
    \includegraphics[width=.93\linewidth]{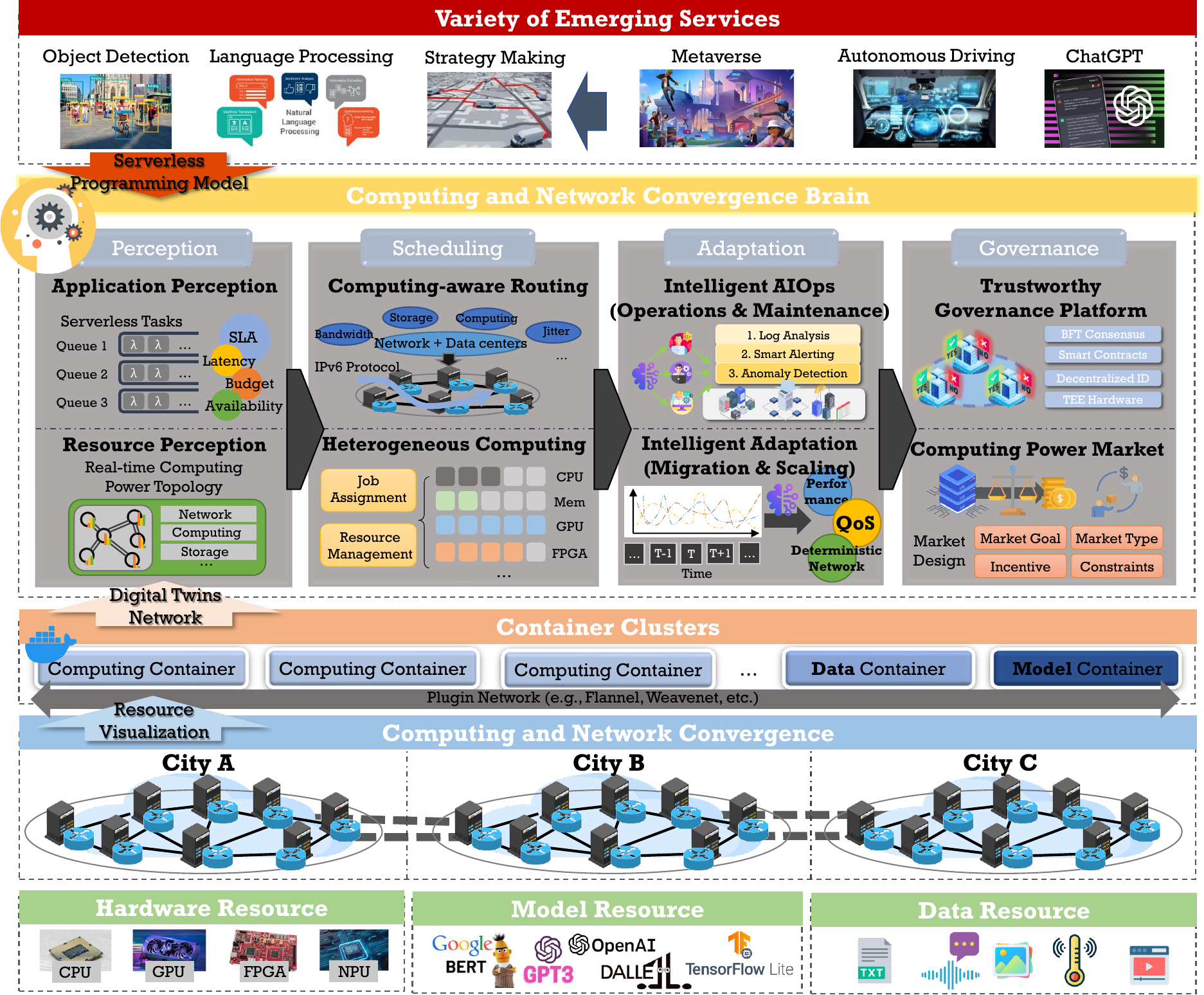}
    \caption{Framework overview of the CNC brain.}
    \label{fig:CNC_overview}
\end{figure*}

To coordinate various resources and computation tasks distributed in the CNC efficiently, we present CNC brain, the concept of an intelligence-endogenous management platform for the CNC.
As shown in Fig. \ref{fig:CNC_overview}, it comprises four essential building blocks described in detail as follows.

\subsection{Perception} 
\label{sec:perception}

As a prerequisite to exploring the potential of the CNC, the perception aims to get a comprehensive understanding of the demand and supply in the CNC. 

In terms of demand perception, besides the service-level objective (SLO) of applications, the CNC brain needs to make in-depth analyses of application characteristics, such as request execution processes and request arrival patterns.

\paragraph{Request Execution Process} Inadequate resource allocation to application instances results in SLO violations. 
Therefore, the CNC brain needs to ascertain the execution process of requests (e.g., execution time) with different resource configurations based on the ergodic method or estimation algorithms. 
However, measuring all possible schemes is both time- and resource-consuming. 
It is because the search space will be extremely huge, considering the large number of heterogeneous and continuously updated applications deployed in the CNC. 

\paragraph{Request Arrival Pattern} To satisfy user SLOs while minimizing resource waste in the long term, the CNC brain should dynamically scale instances based on application request arrival patterns (e.g., diurnal and seasonal periodicity). 
However, accurately predicting the request pattern of an application is challenging.
This is because the request of applications often change suddenly, and the request arrival patterns vary greatly from application to application.
% To this end, we build a lightweight request pattern prediction model. The model consists of two parts, a predictor and a compensator. The predictor is responsible for modelling the periodic and non-periodic factors associated with the time-varying workload. The compensator is responsible for correcting the output of the predictor based on recent prediction errors.

In terms of supply perception in the CNC, it is expected to measure the state of computing nodes, such as resource usage and resource characteristics. 
However, the computing nodes in the CNC are highly heterogeneous in terms of brands, models, capacities, programming models supported, and so on. 
Also, the network topology connecting them is highly dynamic and uncertain. 
Therefore, a unified metric framework is needed for an abstract representation of these computing resources in the CNC brain. 
The framework is expected to translate the demand of each application request into polymorphic resource requests for the CNC.
It provides a standard rule for routing of computing tasks, management of computing nodes, billing of computing power, etc., in the following building blocks. 
Besides the unified metric framework, it is challenging to build a real-time and high-fidelity perception on a large-scale CNC, which requires the CNC brain to decide the update object, period, and approach for the freshness of perceptual information (e.g., Age-of-Information) in the CNC.

\subsection{Scheduling} 

Depending on the perception information, the CNC brain considers each computing node's load and application request's constraints to make scheduling decisions.
In other words, it needs not only to coordinate heterogeneous computing resources in computing nodes in the CNC to execute the computing tasks of requests, but also distribute the computing tasks to the proper set of computing nodes in the CNC. 

\paragraph{Heterogeneous Computing Resource Provision} 
The CNC offers a wide choice of processing platforms for computing tasks. 
For example, GPU data centres excel at performing large numbers of arithmetic operations in parallel, making them suitable for AI workloads that rely heavily on parallelism. 
FPGA data centres inherently provide low and deterministic latency for real-time workloads because FPGAs can bypass internal bus structures, and their processing functionality can be customised at the logic port level. 
Cloud TPU is Google's custom-designed ASIC for neural network inference. 
Traditional data centres with CPUs are still essential for data pre-processing, such as data augmentation operations. 
The CNC brain needs to coordinate heterogeneous data centres according to their advantages and disadvantages.

\paragraph{Computing and Network-aware Task Routing} 
For each computing task of users, the CNC needs to route the task request to single or multiple computing nodes, which are the most appropriate to execute, and then return the execution result to the users. 
The routing should consider both the SLA on task deadlines and the computing resource topology of the CNC. 
However, conventional routing/scheduling protocols used in data communication networks cannot be directly used for the CNC. 
It is because these protocols maximizing the instantaneous throughput do not consider that the consumptive nature of requests has a long-term effect on the resource of computing nodes and the competition among different tasks for communication and computation resources.

\subsection{Adaptation} 

During the scheduling, the complicated, dynamic environment of the CNC requires a design of adaptation to help the CNC brain be resilient to failures for reliable computing services, corresponding to the Operations and Maintenance (O\&M) of the service delivery life cycle in the CNC. 

Due to the increase in scale and complexity of the CNC, it is challenging for O\&M teams to perform daily monitoring and repair operations on every cloud, edge, end node, and network device.
Most existing O\&M mechanisms rely on human experts or rule-based strategies to detect anomalies, which may lead to on-call fatigue, waste of human resources and risk of misjudgment in the CNC.
Thus, to raise efficiency, it is beneficial to develop intelligent software systems to automatically tackle the CNC's O\&M problems, called AIOps for the CNC.
For example, while it is widely acknowledged that device user manuals and log files can help with anomaly detection and resolution, the useful information contained in massive and diverse manuals and logs can be extracted by neural language processing.
Moreover, in the CNC, either computing or networking services can be interrupted due to software and hardware failures. 
To efficiently deliver robust computing and networking services to clients, the CNC brain should be able to lively migrate the services and allocate more backups to mask underlying hardware and software failures. 
% Existing solutions for dynamically scaling computing resources are designed to provision computing resources based on traffic peaks, resulting in low resource utilisation and high computing resource costs. 
% In addition, existing service migration schemes are heuristic schemes designed based on expert experience, they make a one-shot decision and fail to capture delay rewards brought by future computing service requests.

\subsection{Governance} 

Last but not least, given the large number of participants and the pluralistic interests behind them in the CNC, the CNC brain aims to develop a socially oriented governance structure. 
Governance will reap the benefits of the CNC by developing specific rules for computing power. 
Particularly, a comprehensive market for computing power is needed to incentivise trading for highly efficient resource use. 
Also, a self-organising and collectively governed decentralised strategy execution engine is needed to provide governance over the behaviour of resource providers without compromising their control over their own resources.

\paragraph{Computing Power Market} To promote the CNC's sustainable development, computing power's economic value should be explored. However, different from traditional real products in which the value can be easily determined, it is hard to precisely evaluate the quality of computing power in a fair and transparent way due to its inherent characteristics, e.g., heterogeneous hardware, diversified algorithms and even service requirements. 
More importantly, different from the traditional market, the reusability of intermediate computing results and the sharing of computing power (e.g., AI batch inference) will make the market design very complicated.
Thus, the CNC brain should build a computing power market based on the unified metric framework developed in the perception in Sec. \ref{sec:perception} and some features of workloads in the CNC (e.g., shareability and reusability).

\paragraph{Trustworthy Decentralized Governance} The CNC consists of large, distributed and diverse computing nodes and users. 
The fundamental and general principle of good governance for society is that every member can participate in policy making and reach consensus, and that policies should be enforced transparently and without bias against particular users. 
However, due to the CNC's wide distribution and Byzantine environment, malicious attackers can disrupt any phase of governance. 
Achieving social consensus for policy making and enforcement, and resisting malicious behaviour across the CNC is critical for governance. 
Blockchain is an emerging and promising distributed technology for public and tamper-proof consensus, and its smart contract can be used for automated script execution. 
Thus, based on blockchain technology, the CNC brain intends to build a participatory and consensus-oriented decentralised self-governance platform as the infrastructure of CNC governance.

\subsection{Pathway of Implementation}

The CNC brain follows a master-slave architecture. We deploy the proposed CNC brain in a group of delegated servers (e.g., the servers run by the government or enterprise alliances) as \emph{master nodes}. The CNC participants (e.g., application developers, model developers, data vendors, and cloud/edge providers) connect to the master nodes, catalogue available AI resources, and become \emph{worker nodes} in the CNC.
The master nodes take the information from the worker nodes and job descriptions from the users as input and send scheduling, adaptation and governance strategies to the worker nodes.

\section{A Reinforcement Learning-based Intelligent Resource Management for Computing and Network Convergence}

\begin{figure}[t]
    \centering
    \includegraphics[width=\linewidth]{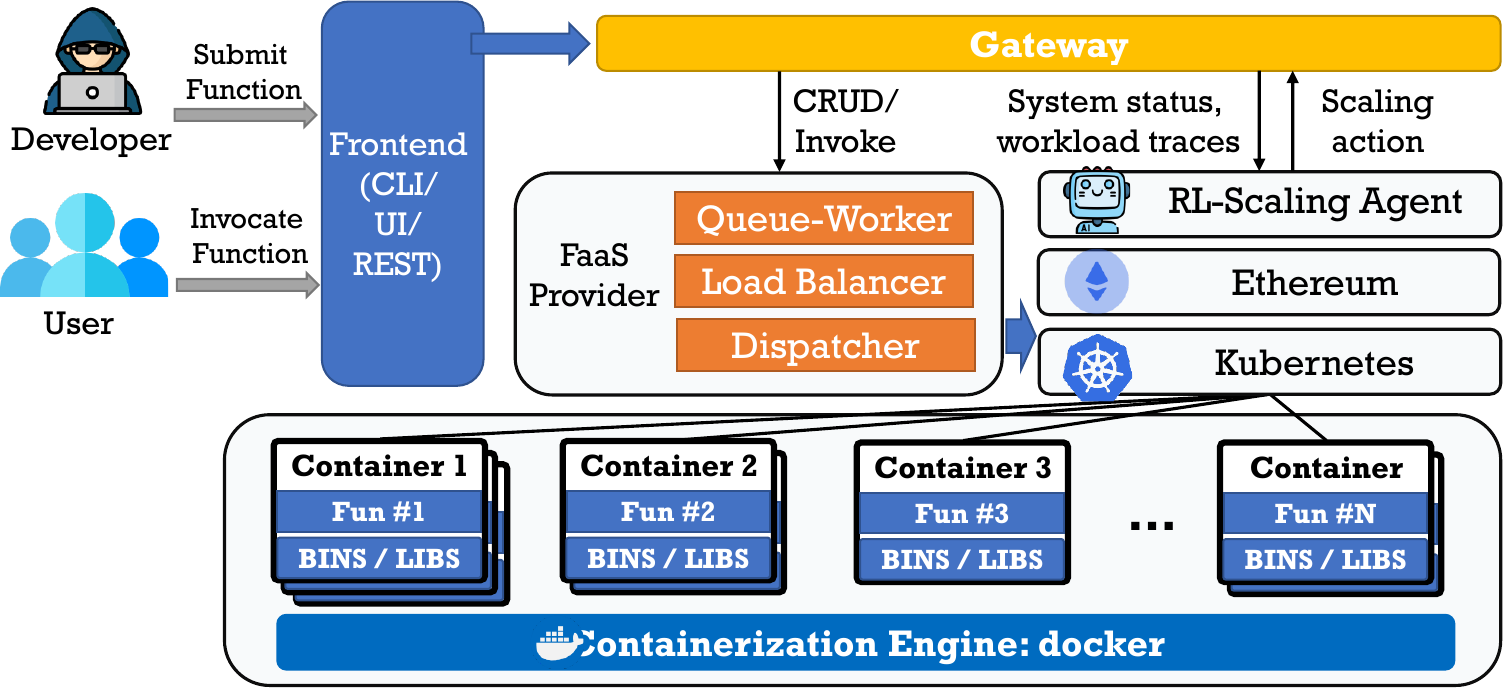}
    \caption{Framework design of DRL intelligent resource management of CNC.}
    \label{fig:rl_framework}
\end{figure}

Fig.~\ref{fig:rl_framework} shows the framework overview of the prototype. 
We implement a prototype of the CNC brain over a serverless platform OpenFaaS\footnote{https://github.com/openfaas/faas}. 
It provides the three most common access methods in the front end, i.e., command line interface, UI interface, and RESTful API. 
With this prototype, developers can easily deploy event-driven functions and run applications without managing backend infrastructure, while users can invoke the deployed functions via an HTTP REST call. 
All requests will be sent to the gateway responsible for authentication and request forwarding. 
The faas-provider is the core module for serverless computing, providing the CRUD capabilities for the deployed functions. 
To achieve this, the faas-provider has a queue-worker for request consumption and asynchronous invocation, a load balancer for efficiently distributing incoming requests across a group of function instances, and a dispatcher for request delivery. 
As a common practice, we run the deployed functions on top of a containerization engine Docker\footnote{https://www.docker.com/} and use a container orchestration platform Kubernetes\footnote{https://github.com/kubernetes/kubernetes} for automating deployment and management of containerized functions.

A DRL agent is deployed in this prototype to achieve resource management and proactive scaling. Specifically, we implement a deep recurrent q-learning-based (DRQN) algorithm~\cite{hausknecht2015deep,mnih2015human} to adjust the number of instances of the summation function dynamically. 
DRQN is a powerful extension of traditional deep q-learning, replacing the fully-connected layer with a recurrent LSTM to investigate the temporal relationships of inputs. 
DRQN makes scaling decisions at regular intervals. 
For each decision-making, DRQN takes the system status and workload traces as input and outputs the scaling decision. 
The input state is a four-dimensional vector, consisting of the number of instances, the average requests per second, the average CPU usage, and the average latency violation rate. 
The output action is a discrete variable ranging from 0 to 4, which means that the number of instances is minus one, unchanged, plus one, plus two, plus four. 
We build a simulated environment based on the historical traces and train the DRQN to accelerate training. 
After convergence, we deploy the trained DRQN to the real environment. 
Moreover, each node joins an Ethereum\footnote{https://ethereum.org} Test Network, which records the DRL agent's resource management and proactive scaling strategy via smart contract for a prototype of trustworthy decentralized governance. The blockchain in our prototype is pluggable, meaning that it can be replaced by other blockchains that support smart contracts but have different consensus protocols (e.g., PoW, PoS, and PBFT) and different permission models (e.g., permissionless and permissioned).

\section{Performance Evaluation}
\label{sec:performance_evaluation}

\subsection{Evaluation Setup}

To verify the feasibility and effectiveness of our proposed framework, we conduct experiments with the following settings. We deploy our prototype to a PC with Intel Core i7-7700HQ CPU@2.8GHz and 16GB of memory. To simulate the traffic of real-world business scenarios, we use a summation function as the workload and trigger it according to the publicly released traces of Microsoft's Azure Functions \cite{254430}, which records the number of invocations per minute for each function. For each instance that actually processes the request (i.e., a pod in Kubernetes), it has a CPU limit of 200 milliCPU. And the maximum number of instances is set to 5. The latency threshold for each request is 2.5 seconds. For the DRQN-based scaling agent, the neural network model is formed by sequentially connecting a fully connected layer of size 4$\times$128, a 2-layer LSTM of size 128$\times$128, and a fully connected layer of size 128$\times$5. The intervals for DRQN decision-making is 15 seconds, which is consistent with the default time interval of OpenFaaS scaling.

By default, OpenFaaS supports auto-scaling by manually-configured alarm rules. For comparison, we use the following alarm rules as baselines:
\begin{itemize}
	\item \textbf{OpenFaaS-RPS$>$5}: This is the default scaling setting of OpenFaaS, which scales up when the requests-per-second is higher than 5. When the alarm is resolved, scale down. 
	\item \textbf{OpenFaaS-RPS$>$2}: This is a conservative strategy that scales up the function when the requests-per-second is higher than 2. When resolved, scale down.
	\item \textbf{OpenFaaS-VPS$>$1}: Scale up when the violations-per-second is higher than 1. When resolved, scale down.
\end{itemize}

\subsection{Evaluation Results}

\begin{figure}[t]
    \centering
    \includegraphics[width=0.85\linewidth]{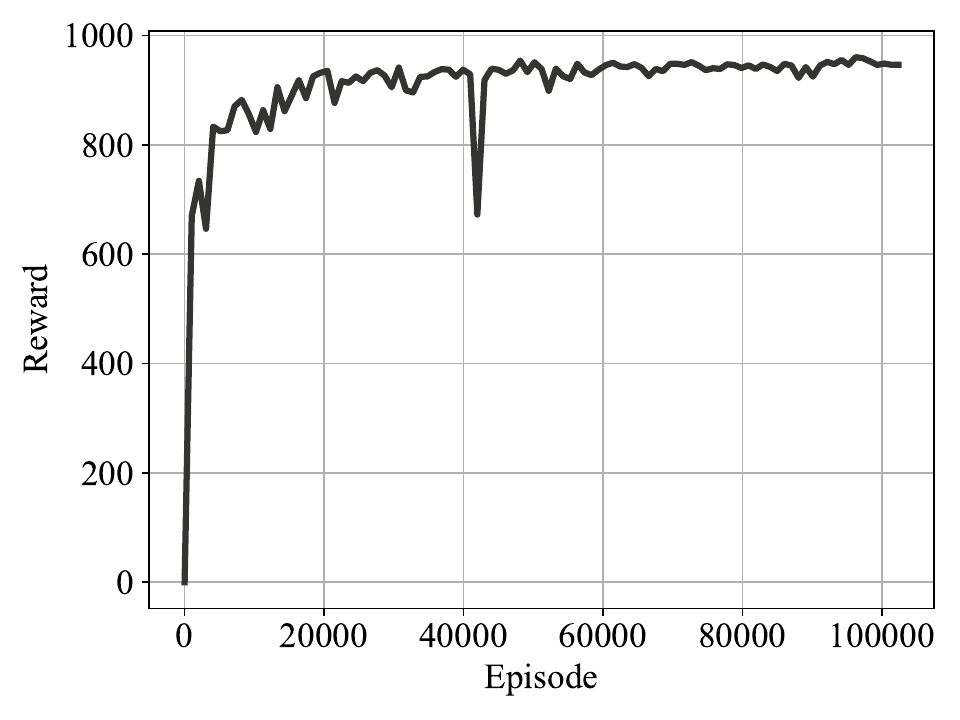}
    \caption{Convergence of reward during the DRL training.}
    \label{fig:convergence}
\end{figure}

% (... the performance of convergence for the proposed scheme)
We first evaluate the convergence performance of the DRQN algorithm and plot the episodic reward in Fig.\ref{fig:convergence}. The steps for each test episode are 1000. For each step, if the number of instances after performing the scaling action can meet the load while maintaining the minimum, the reward for the current step is 1; otherwise, the reward is 0. Thus the maximum episodic reward is 1000. We can see that the episodic reward of DRQN during training gradually converges to a high value, which is around 950. 

\begin{figure}[t] 
	\centering
	\subfigure[Invocation Number vs. Total CPU Seconds]
	{\includegraphics[width=0.85\linewidth]{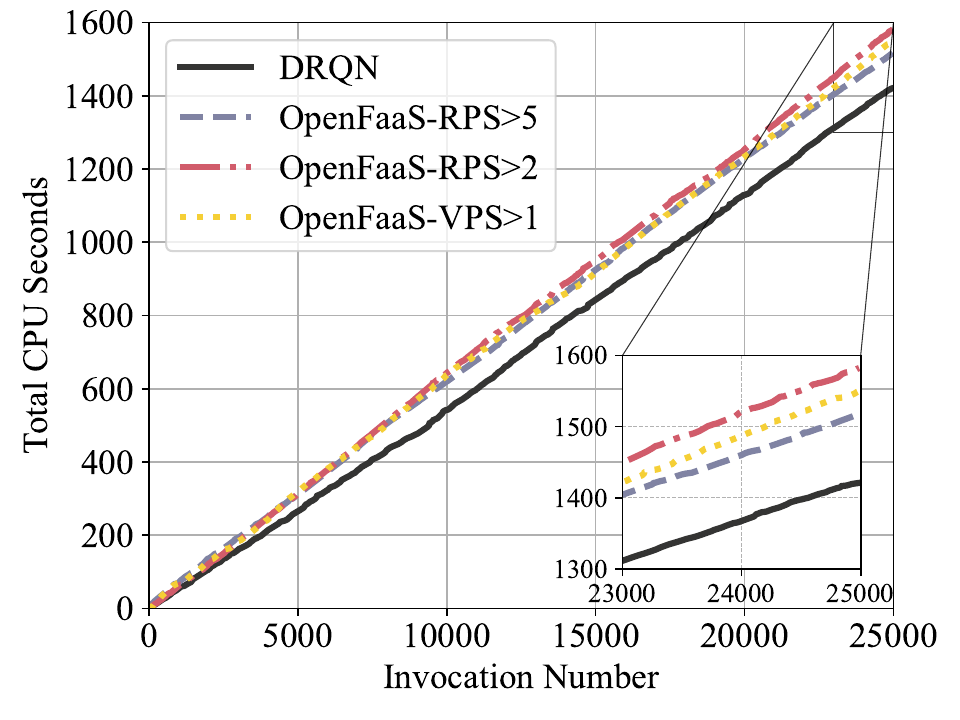}}
	\subfigure[Seconds vs. Average Violation Rate]
	{\includegraphics[width=0.85\linewidth]{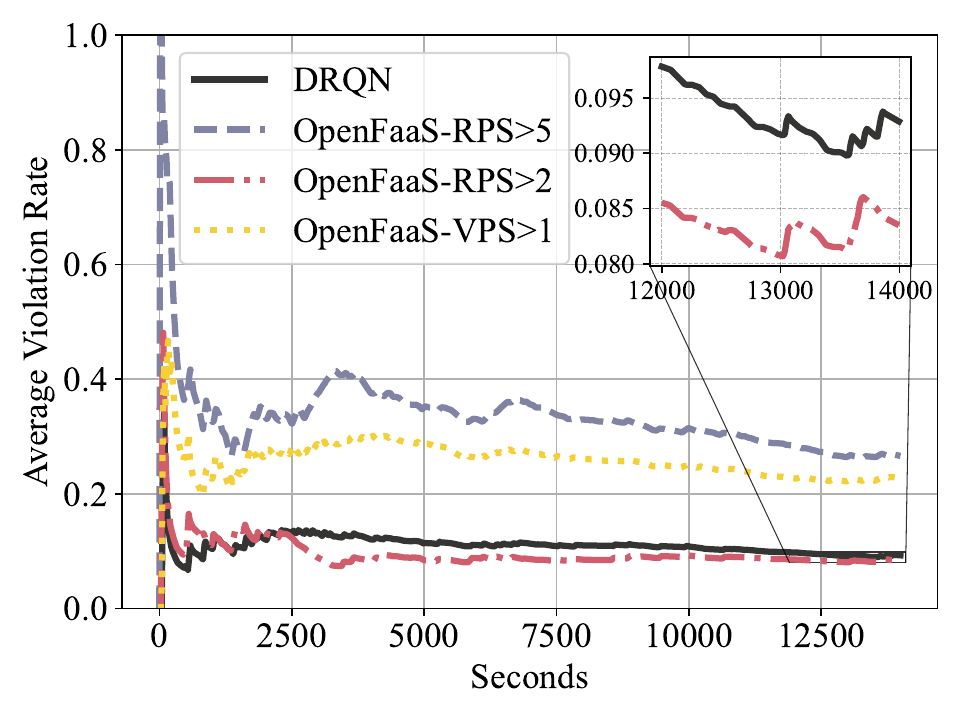}}
	\caption{Performance evaluation of the proposed CNC brain prototype.}
	\label{fig:exp}
\end{figure}

To demonstrate the effectiveness of our algorithm in real environments, we deploy the converged model in our simplified version of the CNC. Auto-scaling aims to reduce the CPU consumption of the system while reducing the latency violation rate. Fig.\ref{fig:exp}(a) first presents the CPU consumption over time. For DRQN and each baseline, we run for approximately four hours. It can be observed that DRQN achieves the lowest CPU consumption for the same number of function invocations. The extra CPU consumption typically comes from two main sources. One is starting too many instances above the need of the requests. Maintaining these instances consumes the CPU. For instance, the conservative scaling strategy OpenFaaS RPS$>$2 has the highest CPU consumption. Second, the number of instances is too low and a large number of requests are sent to the instances (e.g., OpenFaaS RPS$>$5 and OpenFaaS-VPS$>$1). As a result, a large amount of CPU seconds are wasted on the process switching. Compared to the baselines, our prototype can reduce the CPU consumption per request by 6.74\% to 8.69\%.

We further examine our scaling agent's average latency violation rate and  present the results in Fig.\ref{fig:exp}(b). As shown, the violation rate gradually stabilizes as time passes. DRQN has a comparable violation rate to that of the conservative strategy OpenFaaS-RPS$>$2. By contrast, the other two scaling strategies (i.e., OpenFaaS RPS$>$5 and OpenFaaS-VPS$>$1) have much higher violation rates, which are 25.2\% and 22\%, respectively. This is because they are not keenly aware of the increase in load, resulting in an inability to scale in a timely manner. Fig.\ref{fig:exp} indicates the effectiveness of our DRQN-based scaling agent in regulating the number of instances, which can scale up in time when the load increases and scale down when the load decreases.

\section{Conclusion and Future Work}
\label{sec:conclusion}

This article presents our understanding and design of a CNC brain and analyses its challenges.
We also implement a prototype of the CNC brain, and the evaluation results prove its effectiveness. 
Nonetheless, there are several future directions for a CNC brain, which are concluded as follows.

\paragraph{CNC brain for Foundation Model-as-a-Service} AI is undergoing a paradigm shift with the rise of models with large quantities of parameters (e.g., BERT, DALL-E, GPT-3) trained with comprehensive data, which has the potential to be adapted to a wide range of downstream tasks (e.g., language, vision, manipulation). These models are called foundation model (FM).
The capabilities of FMs make them receive widespread attention and able to transform various sectors and industries including law, healthcare, and education. Despite the advantages and potentials, making these large FMs benefit everyone is still challenging because running such large models can be expensive or even infeasible for most users. 
Therefore, a CNC brain is expected to deploy FMs in the CNC and schedule sufficient shared resources to support FMs. Then users can access these powerful FMs in the cloud through their APIs (i.e., foundation model-as-a-service) without concern for resource limitation.
\paragraph{Green CNC brain} Despite the sufficient shared computing resources provided by the computing nodes, CNC brings a lot of electric power consumption and pollution. 
To enjoy the convenience of CNC while pursuing carbon neutrality, a green and environmentally-sustainable CNC is indispensable. 
 Thus, the CNC brain needs to incorporate energy and water usage and carbon footprints into the optimization objective (e.g., power usage effectiveness (PUE), carbon usage effectiveness (CUE) and water usage effectiveness (WUE)) beyond the traditional accuracy or latency performance.
Moreover, the CNC brain should take into account the intelligent management of emerging energy-saving technologies (new chips with energy-saving characteristics and liquid cooling technology in data centres).
% For example, ...
\paragraph{Privacy-preserving CNC brain} During the running of a CNC brain, the information of computing nodes and application requests should be open to the CNC brain, resulting in data and privacy leakage. 
Privacy computing technology can ensure data security when data is available. Common privacy computing technologies include secure multi-party computation and homomorphic encryption. In addition to combining information security technology, the CNC architecture design also requires modifications and consistent standards. Attackers may mainly target multiple CNC layers (e.g., resource abstraction interference) and key data centers for network intrusion, data intrusion, malicious code-infected applications, and unauthorized access. Even the scheduling technique could be a good target if it is a white box or uncertainty-sensitive. Security risk detection, defense, and iterative upgrade are essential in such a highly centralized and cooperative architecture.

\bibliographystyle{IEEEtran}
%\bibliography{raytheon}
\bibliography{Ref}

\end{document}